\documentclass{article}

     \usepackage[preprint]{neurips_2021}

\usepackage[utf8]{inputenc} 
\usepackage[T1]{fontenc}    
\usepackage{hyperref}       
\usepackage{url}            
\usepackage{booktabs}       
\usepackage{amsfonts}       
\usepackage{nicefrac}       
\usepackage{microtype}      
\usepackage{xcolor}         
\usepackage{graphicx}
\usepackage{subfigure}
\usepackage{bbm}
\usepackage{bm} 
\usepackage{xspace}
\usepackage{amsmath}
\usepackage{multirow}
\usepackage{amssymb}
\usepackage{pifont}
\usepackage{subfigure}
\usepackage{threeparttable}
\usepackage{wrapfig}

\title{Multi-dataset Pretraining: A Unified Model for Semantic Segmentation}

\author{%
Bowen Shi$^{1}$ ,  Xiaopeng Zhang$^{2}$ , Haohang Xu$^{1}$, \\ \bf Wenrui  Dai$^{1}$, Junni Zou$^{1}$, Hongkai Xiong$^{1}$, Qi Tian$^{2}$\\
$^1$Shanghai Jiao Tong University \; $^2$Huawei Inc. \\
\texttt{\small\{sjtu\_shibowen, xuhaohang, daiwenrui, zoujunni, xionghongkai\}@sjtu.edu.cn}\\
\texttt{\small zxphistory@gmail.com} \; \texttt{\small tian.qi1@huawei.com}
}

\begin{document}
\maketitle
\begin{abstract}
 Collecting annotated data for semantic segmentation is time-consuming and hard to scale up. In this paper, we \emph{for the first time} propose a unified framework, termed as \textbf{M}ulti-\textbf{D}ataset \textbf{P}retraining, to take full advantage of the fragmented annotations of different datasets. The highlight is that the annotations from different domains can be efficiently reused and consistently boost performance for each specific domain. This is achieved by first pretraining the network via the proposed pixel-to-prototype contrastive loss over multiple datasets regardless of their taxonomy labels, and followed by fine-tuning the pretrained model over specific dataset as usual. In order to better model the relationship among images and classes from different datasets, we extend the pixel level embeddings via cross dataset mixing and propose a pixel-to-class sparse coding strategy that explicitly models the pixel-class similarity over the manifold embedding space. In this way, we are able to increase intra-class compactness and inter-class separability, as well as considering inter-class similarity across different datasets for better transferability. Experiments conducted on several benchmarks demonstrate its superior performance. Notably, MDP consistently outperforms the pretrained models over ImageNet by a considerable margin, while only using less than 10\% samples for pretraining.
\end{abstract}

\section{Introduction}
As a basic computer vision task, semantic segmentation has experienced remarkable progress over the past decades, mainly benefiting from the growth of the available annotations. However, annotating images at pixel level granularity is time-consuming and difficult to scale up. In order to alleviate the dense annotation requirement, semantic segmentation is usually fine-tuned based on a pretrained model, \emph{e.g.,} training on a large-scale ImageNet classification dataset \cite{russakovsky2015imagenet}. While ImageNet pretraining can \emph{de facto} lead to significant performance gains, it suffers from task gap that the pretraining is based on global classification while the downstream task is for local pixel level prediction. In this paper, we arise a critical issue, can we solve the task gap via making use of the available annotations off-the-shelf from diverse segmentation datasets for better performance?

\begin{figure*}
\centering
    \subfigure[Standard pipeline]
    {
    \begin{minipage}{0.3\linewidth}
    \centering
        \includegraphics[width=\linewidth]{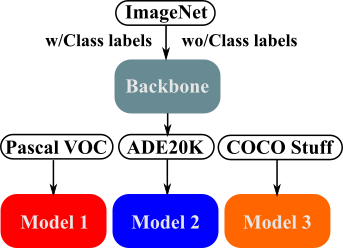}
        \label{fig:imagepretrain}
    \end{minipage}
    }
    \subfigure[Multi-dataset training]
    {
    \begin{minipage}{0.3\linewidth}
    \centering
        \includegraphics[width=\linewidth]{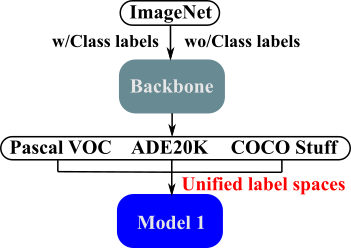}
        \label{fig:multitrain}
    \end{minipage}
    }
    \subfigure[Multi-dataset pretraining]
    {
    \begin{minipage}{0.3\linewidth}
    \centering
        \includegraphics[width=\linewidth]{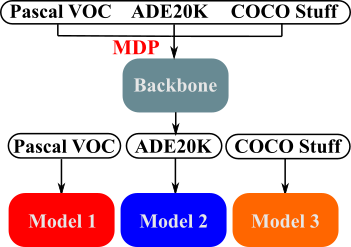}
        \label{fig:ours}
    \end{minipage}
    }
    \caption{Typical settings for semantic segmentation. a) Most of the previous works follow the standard pipeline, which performs supervised or self-supervised pretraining on ImageNet, and then trains segmentation models on one specific dataset alone. b) Multi-dataset training means training a unified segmentation model using all annotated data, but needs manually unifying the label space of different datasets. c) Our work directly uses multiple datasets for pretraining, allowing the model to learn pixel level distinguishable features automatically.}
    \label{main}
\end{figure*}

Though promising it is, a major challenge of unifying multi-datasets for training is to tackle the label inconsistency, where taxonomy from different datasets differs, ranging from class definition and class granularity. For example, the classes \textit{'wall-brick'}, \textit{'wall-concrete'} and \textit{'wall-panel'} in COCO-Stuff \cite{caesar2018coco} are simply labeled as \textit{'wall'} in ADE20K \cite{zhou2018semantic}, and as \textit{'background'} in Pascal VOC \cite{pascalvoc}. However, integrating different datasets into a common taxonomy is time-consuming and error-prone. In this paper, we propose a novel training framework that is able to directly unify different datasets for training regardless of its taxonomy labels. In Fig. \ref{main}, we illustrate several typical pretraining settings for semantic segmentation. The advantage of MDP is that we are able to conveniently combine multiple datasets for jointly training without any human intervention, and the pretrained model can be used as a backbone for downstream fine-tuning as usual. 

Towards this goal, we rely on contrastive loss that is widely used in self-supervised learning \cite{chen2020simple,he2020momentum} for pretraining. In particular, we adjust the global contrastive loss to a supervised pixel level loss, and construct a pixel-to-class prototype mapping, such that the pixel embeddings with the same label enjoy better intra-class compactness and inter-class separability, which we find is beneficial for semantic segmentation task. Considering that classes from different datasets may share similar embeddings, to better modeling inter-class relationship defined by the provided labels, we enrich the pixel level embeddings via cross dataset mixing and extend the pixel-to-class mapping, which can be treated as hard encoding using the provided annotations to sparse coding, which considers the relationship of similar classes. In this way, the pixel level embedding is endowed with a more smooth representation, which is beneficial for better transferability. Experiments conducted on several benchmarks demonstrate its superior performance.

In a nutshell, this paper makes the following contributions:

\begin{itemize}
    \item We propose a novel pretraining setting that is able to effectively make use of the available annotations off-the-shelf by unifying multiple datasets without considering label integration. As far as we know, we are the first to unify multiple labeled datasets for pretraining, while not be bothered by the chaotic labels from diverse datasets. 
    \item We propose a pixel-to-prototype mapping to effectively model intra-class compactness and inter-class separability. To better make use of cross dataset samples, we enrich the pixel embeddings via cross dataset mixing and extend the pixel-to-class hard coding to cross-class mapping via pixel-to-class sparse coding.
    \item MDP surpasses ImageNet pretraining by a large margin, with less than $10\%$ samples comparing with models pretrained over ImageNet. Our findings validate that it is unnecessary for semantic segmentation to make use of classification as a pretext task, this may shed light on future research direction with respect to designing an appropriate pretrained model for semantic segmentation.
\end{itemize}

\section{Related Work}

\textbf{Contrastive learning.} 
Contrastive learning-based methods learn representations by contrasting positive pairs against negative pairs in a discriminative fashion. Recent works mainly benefits from instance discrimination \cite{wu2018unsupervised}, which regards each image and its augmentations as one separate class and others are negatives \cite{he2020momentum,chen2020simple,dosovitskiy2015discriminative,chen2020improved, hjelm2018learning, oord2018representation, tian2019contrastive, grill2020bootstrap}. 
Since using a large number of negatives is crucial for the success of contrastive loss-based representation learning, \cite{wu2018unsupervised} uses a memory bank to store the pre-computed representations from which positive examples are retrieved given a query. Based on it, \cite{he2020momentum} uses a momentum update mechanism to maintain a long queue of negative examples for contrastive learning. 

Thanks to recent explorations in supervised contrastive learning \cite{supcon} and pixel level contrastive learning \cite{xie2020propagate}, some recent works introduce contrast learning into semantic segmentation tasks \cite{efficientlabel,liu2021boot,inigo2021semi,wang2021explore,wounter2021unsupervised} by pulling close the embedding of pixels with the same label and pushing apart the embedding of pixels with different labels. However, these methods do not explore the wide applicability of pixel level contrastive learning, and lack innovation in the pixel level loss design.

\textbf{Multiple datasets training.} 
For recognition tasks like object detection and semantic segmentation, training on naively combined datasets yields low accuracy
and poor generalization \cite{lambert2020mseg} since different datasets have different class definitions and class granularity. Dataset unification, which involves merging different semantic concepts, is important for multi-dataset training. \cite{liang2018dynamic} manually builds a semantic concept hierarchy by combining labels from all four popular datasets and explicitly incorporates the hierarchy into network construction. \cite{lambert2020mseg} manually unifies the taxonomies of 7 semantic segmentation datasets and uses Amazon Mechanical Turk to resolve inconsistent annotations between datasets. However, these methods need heavily manual effort. \cite{zhao2020object} and \cite{zhou2021simple} trains a universal detector by first training a single partitioned detector on multiple datasets with shared backbone dataset-specific outputs, and loss and then unifies the outputs of the partitioned detector in a common taxonomy.
These methods still rely on partitioned learning on their respective datasets. Unlike these works, we do not unify their label space but make the pixel level features distinguishable by multiple datasets pretraining.

\section{Method}
\begin{figure*}[t]
  \centering
  \includegraphics[width=0.85\textwidth]{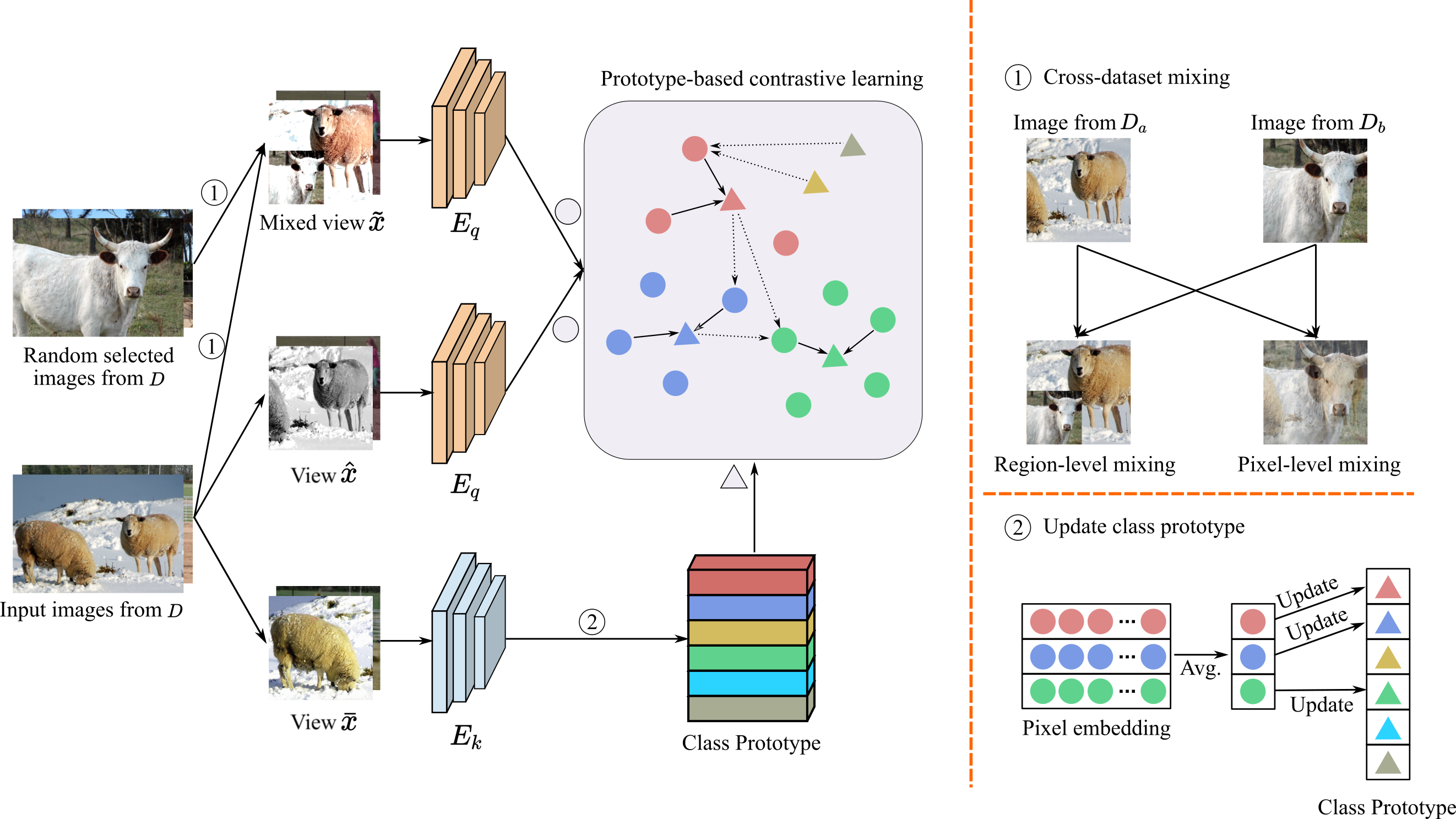}
  \vspace{-0.2cm}
  \caption{The overall pipeline of MDP. Given labeled image $x$ randomly sampled from a collection of multiple datasets $\mathcal{D}$, we first obtain two different views $\hat{x}$ and $\bar{x}$ through data augmentation as well as a mixed view $\tilde{x}$ through cross-dataset mixing with a sample that is also randomly selected from $\mathcal{D}$. Then we conduct class-to-prototype sparse coding to model intra-class compactness and inter-class separability, as well as considering cross-class similarity. The right part illustrates two components of MDP, \emph{i.e.,} cross-dataset mixing and online class prototype update.}
  \label{pipline}
\end{figure*}

In this section, we elaborate on our proposed multi-dataset pretraining strategy. The whole procedure is shown in Fig. \ref{pipline}. Given a set of samples with pixel level annotations, we aim at constructing pixel-to-class prototype mapping to model intra-class compactness and inter-class separability. The core modules consist of two parts, 1) pixel-to-prototype contrastive loss and 2) cross dataset learning, which would be explained in detail in the following sections. 

\subsection{Preliminaries}
Contrastive learning targets at training an encoder to map positive pairs to similar representations while pushing away the negative samples in the embedding space. Given unlabeled training set $\bm{X}=\left \{ x_{1},x_{2},...,x_{n} \right \}$. Instance-wise contrastive learning aims to learn an encoder $E_{q}$ that maps the samples $\bm{X}$ to embedding space $\bm{V}=\left \{ v_{1},v_{2},...,v_{n} \right \}$ by optimizing a contrastive loss. Take the Noise Contrastive Estimator (NCE) \cite{oord2018representation} as an example, the contrastive loss is defined as:

\begin{equation}
  \label{InfoNCE}
  \mathcal{L}_{nce}(x_{i},{x}'_{i}) =  -\mathrm{log}\frac{\mathrm{exp}(E_{q}(x_{i})\cdot E_{k}({x}'_{i})/\tau )}{\mathrm{exp}(E_{q}(x_{i})\cdot E_{k}({x}'_{i})/\tau )+\sum_{j=1}^{K}\mathrm{exp}(E_{q}(x_{i})\cdot E_{k}({x}'_{j})/\tau ))},
\end{equation}

where $\tau$ is the temperature parameter, and ${x}'_{i}$ and ${x}'_{j}$ denote the positive and negative samples of ${x}_{i}$, respectively. The encoder $E_{k}$ can be shared \cite{chen2020simple,caron2020unsupervised} or momentum update of the encoder $E_{q}$~\cite{he2020momentum}.

\subsection{Pixel-to-pixel Contrastive Learning} 
\label{sec:baseline}
Inspired by contrastive learning, we first present a simple baseline that directly extends the instance level contrastive learning to pixel level, guided by the provided pixel level annotations. In this setting, pixels with the same label are treated as positive pairs and pulled together, while those with different labels are regarded as negative pairs and pushed away. Note that since we constrain the positive samples within the same label, making it conveniently applicable for multiple datasets.

Specifically, given multiple labeled datasets $\mathcal{D}=\{\mathcal{D}_1,\mathcal{D}_2,...,\mathcal{D}_N\}$ along with label space $\mathcal{Y}=\{\mathcal{Y}_1,\mathcal{Y}_2,...,\mathcal{Y}_N\}$, we randomly select $n$ samples $\{x_t\}_{t=1}^n$ from $\mathcal{D}$ regardless of which domain each sample comes from. Denote $\hat{x_t}$ and $\bar{x_t}$ be two different augmented views of image $x_t$, with ground truth pixel level class label map $\hat{Y}=[\hat{y}_i \in \mathcal{Y}]$ and $\bar{Y}=[\bar{y_i} \in \mathcal{Y}]$, where $i$ denotes pixel in the image.
$\hat{x}_t$ and $\bar{x}_t$ are separately sent to feature extractor $E_q$ and $E_k$ to obtain $d$-dimensional per-pixel unit-normalized features $\hat{F}=[\hat{f}_i]$ and $\bar{F}=[\bar{f}_i]$, where $E_k$ is a momentum update version of $E_q$. For pixel $i$ in $\hat{F}$, the pixel $j$ in $\bar{F}$ with the same class label is considered as positive sample of $i$, and the pixel level contrastive loss is computed by:

\begin{equation}
\mathcal{L}_{pixel}= - \frac{1}{\bar{N}_{y_{i}}} \sum_{j=1}^{N} \mathbbm{1}\left[\hat{y}_{i}=\bar{y_{j}}\right] \log \left(\frac{\exp \left(\hat{f}_{i} \cdot \bar{f}_{j} / \tau\right)}{\sum_{k=1}^{N} \exp \left(\hat{f}_{i} \cdot \bar{f}_{k} / \tau\right)}\right),
\end{equation}

where $N=\hat{N}=\bar{N}$ denotes the total number of pixels in each view and $\bar{N}_{y}$ denotes the number of pixels with class $y$ in $\bar{x}_{t}$. The loss is averaged over all pixels on the first view. Similarly, the contrastive loss for pixel $j$ on the second view is also computed and averaged.

The pixel level contrastive loss enables multi-dataset training and does not need to consider the label mapping, since the calculation of pixel level contrastive loss is independent for each image. However, such a pixel-to-pixel optimization strategy is sensitive to noisy annotations and more importantly, it does not consider the relationship across datasets, which limits its representation ability. In the following, we extend this baseline to support modeling relationships among different images and different classes for better pretraining. 

\subsection{Pixel-to-prototype Contrastive Learning}
\label{sec:prototype}
Considering that the pixel level representation is less meaningful than the instance-level one, and may suffer great difference even for embeddings with the same label, such representation uncertainty makes the pixel-to-pixel optimization goal fluctuate and hard for convergence. To solve this issue, we adjust the pixel-to-pixel contrastive learning with more robust pixel-to-class mapping. The motivation is that class-level representation is more stable and memory-efficient comparing with recording pixel level embeddings. In particular, we merge the label spaces of the collection of all datasets $\mathcal{D}$ and obtain class set $\mathcal{Y}=\mathcal{Y}_{1} \cup \mathcal{Y}_{2} \cup ...\cup \mathcal{Y}_n$. We maintain a prototype for each class $y_i \in \mathcal{Y}$ in the memory bank. Then we learn the embeddings of all the pixels in each input image $x$ by pulling them close to the same class prototype and pushing them apart to different class prototypes.

\paragraph{Class prototype.} 
A key component for pixel-to-prototype mapping is to maintain the prototype for each class. Considering that the labeling granularity is at pixel level, which is huge and memory-consuming, we propose an efficient prototype maintenance strategy to dynamically update the class embeddings. Specifically, suppose we have a total of $N$ training images and $|\mathcal{Y}|$ semantic classes. Let $P_y$ be the class prototype of class $y$, we first calculate the embedding of class $y$ for $n$-th images, represented by $p_{yn}$, by average pooling of all the embeddings of pixels labeled as $y$ in the $n$-th image. Then, $P_y$ is obtained by averaging all embeddings among $N$ images that contain class $y$:
\begin{equation}
P_y = \frac{\sum_{i=1}^{N}m_{yi}p_{yi}}{\sum_{i=1}^{N}m_{yi}},
\label{eq2}
\end{equation}
where $m_{yi} \in [0,1]$ is a binary mask indicating whether the $i$th-image contains pixels with class $y$. In practice, in order to realize the dynamic update of $P_y$, we store the embedding sum of class $c$ and the number of images that contain pixels with class $c$ for each batch $b$.
Some works \cite{wang2021explore,wounter2021unsupervised} propose region embedding strategy, which simply stores the image embedding set $P'_{y} = \{p_{y1},p_{y2},...,p_{yN}\}$ for class $y$ in the memory bank. We argue that our two step average strategy for prototype calculation can bring three advantages comparing with region embedding: 

$\bullet$ \textbf{Reducing the storage requirements}. The memory bank of region embedding is built with size $|\mathcal{Y}| \times N_b \times Dim$, where $Dim$ indicates the dimension of pixel embedding and $N_b$ indicates the number of image embeddings saved in the memory bank. Its storage consumption will become large as the growth of $N_b$ and $|\mathcal{Y}|$. While using the class prototype, we can reduce the size to $|\mathcal{Y}| \times B \times Dim$ while maintaining the dynamic update of $P_y$, where $B = N_b/bs$ is the total number of batches and $bs$ means the batch size.

$\bullet$ \textbf{Alleviating the class imbalance problem especially when using multiple datasets}. Our method obtains one prototype for each class $y$, regardless of the number of images containing $y$. This ensures the contribution of small datasets and classes with rare data be not ignored. 

$\bullet$ \textbf{Obtaining more robust class embeddings}. The prototypes we provide for each class are not susceptible to dirty data and are representative enough since it is obtained by considering the embeddings of all pixels with the same class. 

\paragraph{Loss function.} In prototype-based contrastive learning, 
we first update the prototype in the memory bank using $\bar{F}$ according to (\ref{eq2}). Then, for pixel $i$ with class $j$ in $\hat{x}$, we maximize the agreement between its embedding $\hat{f}_i$ and the prototype of class $j$. Additionally, we require a push-force to avoid collapse in the embedding space. This can be achieved by pushing $\hat{f}_i$ and other class prototypes apart. The prototype-based contrastive loss for pixel $i$ is computed by: 
\begin{equation}
\mathcal{L}_{class} = - \mathbbm{1}\left[\hat{y}_i=j\right] \log \left(\frac{\exp \left(\hat{f}_{i} \cdot P_{j} / \tau\right)}{\sum_{k=1}^{|L|} \exp \left(\hat{f}_{i} \cdot P_{k} / \tau\right)}\right)
\end{equation}
The final loss is also computed for pixels from two different views and then averaged. Pixel embeddings in current image are not only influenced by the same class pixels in other images, but also interactive with the classes which do not occur in current image. Prototype-based contrastive learning makes the learning of embedding not confined to a single image, which greatly improves the embedding quality compared to pixel-to-pixel contrastive learning.

\subsection{Cross Dataset Learning}
The pixel-to-prototype mapping can be treated as hard coding for each pixel, and the codes are simply generated by averaging operations using the provided pixel level labels. While in real applications, classes from different datasets may share similar embeddings. In order to better model the inter-class relationship defined by the provided labels, we propose two cross dataset interaction operations. First, we enrich the pixel level embeddings via cross dataset mixing. And second, we extend the pixel-to-class hard mapping to more general soft mapping. In this way, the pixel level embedding is endowed with a more smooth representation, which is beneficial for better transferability.

\paragraph{Cross-image pixel representation.}

We adopt two different levels of data mixing methods, \emph{i.e.,} region-level mixing and pixel-level mixing, to interact representation across different datasets. We argue that data mixing is intrinsically suitable for pixel level tasks, though it is first proposed for improving classification on ImageNet \cite{russakovsky2015imagenet}. These two mixing methods alleviate the problem of input inconsistency and class inconsistency from different perspectives, and they are complementary.

\emph{Region-level mixing} Given two labeled images \{$x_i$,$y_i$\},\{$x_j,y_j$\} from the collection of all datasets  $\mathcal{D}$, we conduct cross-dataset cutmix \cite{yun2019cutmix} operation by:
\begin{equation}
\begin{array}{l}
\tilde{x}=M \odot x_i +(1-M) \odot y_j \\
\tilde{y}=M \odot y_i+(1-M) \odot y_j
\end{array}
\end{equation}
where $M$ denotes a binary mask indicating where to dropout and fill in from two images, and $\odot$ is element-wise multiplication. The mixed image can interact with the prototype in the same way as before. Region-level mixing enables the model to see different regions of different datasets at the same time, which can reduce the gap between different datasets. At the same time, region-level mixing destroys the regional continuity of the original image, making the model pay more attention to pixel level details and also increase the diversity of features.

\emph{Pixel-level mixing} Also, given two random samples \{$x_i$,$y_i$\},\{$x_j,y_j$\}, we can conduct cross-dataset pixel-level mixing using mixup \cite{zhang2017mixup} operation:
\begin{equation}
\begin{array}{l}
\tilde{x}= \lambda x_i+(1-\lambda)x_j\\
\tilde{y}= \{\lambda,y_i,y_j\}
\end{array}
\end{equation}
Where $\lambda \in[0,1]$. It should be noted that the class label should be an integer, so we do not mix the label but storing the set $\{\lambda,y_i,y_j\}$ in $\tilde{y}$. the final loss $\mathcal{L}(\tilde{x},\tilde{y},P)$ can be obtained by considering the contribution of the two groundtruth labels, $y_i$ and $y_j$, using the class prototype $P$ and weight $\lambda$:
\begin{equation}
\mathcal{L}(\tilde{x},\tilde{y},P)= \lambda \mathcal{L}(\tilde{x},y_i,P) + (1-\lambda)\mathcal{L}(\tilde{x},y_j,P).
\end{equation}
Pixel-level mixing can integrate the image information from different datasets in a more fine-grained way. In addition, pixels will contain contents from different classes, which makes up for inconsistencies across datasets at both image and class levels.

\paragraph{Cross-class sparse coding.}

We explicitly model the inter-class similarity via extending the pixel-to-prototype hard coding to soft coding, \emph{i.e.,} in addition to pushing the embedding of pixel $i$ close to its corresponding ground-truth class prototype, we also push its embedding close to its top-$K$ similar class prototypes:
\begin{equation}
\scriptsize
\mathcal{L}_{sc} = - \alpha * \mathbbm{1}\left[\hat{y}_{i}=j\right] \log \left(\frac{\exp \left(\hat{f}_{i} \cdot P_{j} / \tau\right)}{\sum_{k=1}^{|L|} \exp \left(\hat{f}_{i} \cdot P_{k} / \tau\right)}\right) - (1-\alpha) \frac{1}{K} \sum \mathbbm{1}\left[t 
\in T\right] \log \left(\frac{\exp \left(\hat{f}_i \cdot P_{t} / \tau\right)}{\sum_{k=1}^{|L|} \exp \left(\hat{f}_i \cdot P_{k} / \tau\right)}\right),
\end{equation}
where T denotes the set of top-$K$ similar class prototypes of $i$. $\alpha$ is simply set to $0.5$ and $K$ is $5$ by default. Our method provides an elegant way to utilize the class similarity of different datasets to help improve the diversity of pixel features.

\section{Experiments}
\label{sec:exp}
In this section, we evaluate MDP on several widely used benchmarks, as well as a detailed ablation study to reveal how each module affects the final performance.

\subsection{Experimental Setups}

\textbf{Datasets.}
Our experiments are conducted on four datasets, namely:
\begin{itemize}
    \item  \textbf{Cityscapes} \cite{cityscapes} has 5,000 finely annotated urban scene images, with 2,975/500/1,524 for train/val/test, respectively. The segmentation performance is reported on 19 challenging categories, such as person, sky, car, and building \emph{etc}.
    \item \textbf{Pascal VOC 2012} \cite{pascalvoc} consists of 10,582 training (including the annotations provided by \cite{pascalaug}), 1,449
validation, and 456 test images with pixel level annotations for 20 foreground object classes and one background class.
    \item \textbf{ADE20K} \cite{zhou2018semantic} 
    contains around 25K images spanning 150 semantic categories, of which 20K for training, 2K for validation, and another 3K for testing.
    \item \textbf{COCO-Stuff} \cite{caesar2018coco} is a large scale dataset, which includes 164K images from COCO 2017 \cite{lin2014microsoft}. Among them, 118k images are used for training and 5k images are used for validation. It provides rich annotations for 80 object classes and 91 stuff classes. 
\end{itemize}

\paragraph{Evaluation.} We choose the training split of Pascal VOC and ADE20K for pretraining since they are comparable at scale, and evaluate the performance on the two datasets to validate how pretraining using other datasets boosts the performance. We also transfer the pretrained model to Cityscapes, where the model does not see during the pretraining stage to validate its generalization ability. Finally, we add COCO-Stuff for pretraining to validate how large scale dataset helps for pretraining. Following the standard setting, we choose mean Intersection-over-Union (mIoU) for performance evaluation.

\textbf{Implementation details.}
We choose DeepLab-v3 \cite{deeplabv3p} based on a standard ResNet-50 as backbone \cite{he2016deep}. Following the settings in MoCo-v2 \cite{chen2020improved}, we employ an asynchronously updated key encoder and adds a 4-layer projection head after the ASPP layer of the DeepLab-v3 model, which finally results in a 256-d embedding vector for each pixel. 
The model is pretrained using an SGD optimizer with momentum $0.9$ and weight decay $4e\!-\!5$. The batch size and the initial learning rate are set to $32$ and $0.2$ respectively, over $2$ NVIDIA Tesla V100 GPUs. The learning rate is decayed to $0$ by cosine scheduler \cite{ilya2016sgdr}.
The input size is set to $224 \times 224$ for efficiency. We use the same set of augmentations as in MoCo-v2. The temperature parameter $\tau$ of contrastive loss is set to $0.07$, and the size of the memory bank is approximately equal to the total number of batches.

During the downstream fine-tuning stage, we follow the basic configure of MMSegmentation \footnote{\url{https://github.com/open-mmlab/mmsegmentation}} except substituting the backbone with a standard ResNet-50 and removing the auxiliary head. Other settings follow the default configurations as in MMSegmentation. For Pascal VOC, we fine-tune the pretrained model for $40k$ iterations using $513 \times 513$ input size, while for ADE20K, the iterations are set to $80k$ with $512 \times 512$ input size, and for Cityscapes, the iterations are set as $40k$ with $512 \times 1024$ input size. \emph{Note that our implementation aims to show the effectiveness of multi-dataset pretraining and use some basic settings and structure for evaluation}. While there are several tricks to further improve the performance, such as larger resolution for pretraining and more advanced structure with auxiliary head, this is out of the scope of this paper.

\begin{table}[t]

\renewcommand\arraystretch{1.25}
\centering
\caption{
Overall results evaluated on three different datasets: Pascal VOC, ADE20K and Cityscapes, using different amount of datasets for pretraining.}
\label{tm} 
    \begin{tabular}{@{} c |c c c c |c|c c c @{}}
      \toprule
    Method  & \multicolumn{4}{c|}{Pretrained Dataset} &Epoch  & \multicolumn{3}{c}{mIoU}   \\
       & ImageNet & VOC & ADE20K& COCO & & VOC & ADE20K & Cityscpes \\
      \hline
      Scratch & & & & &- & 44.78& 28.67& 54.27\\
      \hline
      MoCo-v2& \checkmark & & & &800 &71.59 &38.29 &77.52\\
      Supervised & \checkmark & & & &-&75.63 &39.36 &77.60 \\
      \hline
      MDP & & \checkmark & \checkmark & & 100& 73.32&40.17 &77.75\\
      MDP & & \checkmark & \checkmark & & 200& 74.30& 40.83&78.59\\
      MDP & & \checkmark & \checkmark &\checkmark& 200&\textbf{77.79} &\textbf{42.69} &\textbf{80.64}\\
      \bottomrule
    \end{tabular}
\end{table}

\subsection{Results}
In this section, we report the overall results when fine-tuning on three datasets: Pascal VOC, ADE20K, and Cityscapes. As shown in Table \ref{tm}, for completeness, we also list the results of using other pretrained models, which includes: a) train from scratch with random initialization; b) self-supervised model based on MoCo-v2 using ImageNet as pretraining data; c) fully supervised model on ImageNet. We report the results for pretraining over both 100 and 200 epochs. From this table we find that:

\textbf{Comparing with other pretrained models.} 
All pretraining models surpass train from scratch by a large margin, and MDP beats both supervised and self-supervised ImageNet pretrained models. Specifically, when using only Pascal VOC and ADE20K (around 30K images) for pretraining, we achieve 74.30\% and 40.83\% accuracies on Pascal VOC and ADE20K, respectively.  The performance can be further improved by adding a large scale COCO Stuff dataset, which surpasses the fully supervised model by a noticeable margin, that is  2.16\% performance gain ($75.63\% \to 77.79 \%$) on Pascal VOC, 3.33\% gain ($39.36\% \to 42.69 \%$) on ADE20K. It should be noted that compared to ImageNet pretraining, MDP only uses less than 10\% samples for pretraining, which is more efficient.

\textbf{Transfer ability.} We also evaluate the performance on Cityscapes, where the model does not see any images over this dataset during pretraining, and the results are surprisingly promising. When only using Pascal VOC and AED20K for pretraining, we achieve 78.59\% accuracy, which is already 1\% better than the fully supervised model on ImageNet, and the performance can be further boosted to 80.64\% when adding COCO-Stuff for pretraining. The results indicate that MDP also enjoys better transferability.

\begin{figure*}
\centering
    \subfigure[]
    {
    \begin{minipage}{0.3\linewidth}
    \centering
        \includegraphics[width=\linewidth]{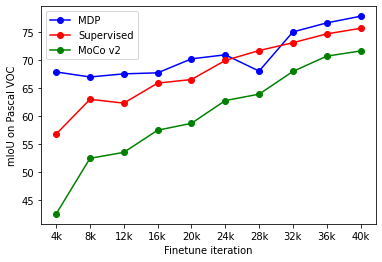}
        \label{fig:main_voc}
        \vspace{-0.3cm}
    \end{minipage}
    }
    \subfigure[]
    {
    \begin{minipage}{0.3\linewidth}
    \centering
        \includegraphics[width=\linewidth]{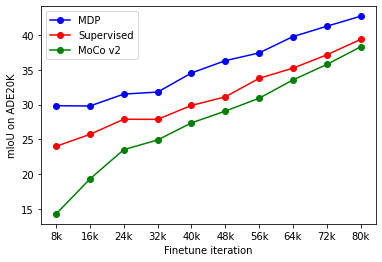}
        \label{fig:main_ade20l}
        \vspace{-0.3cm}
    \end{minipage}
    }
    \subfigure[]
    {
    \begin{minipage}{0.3\linewidth}
    \centering
        \includegraphics[width=\linewidth]{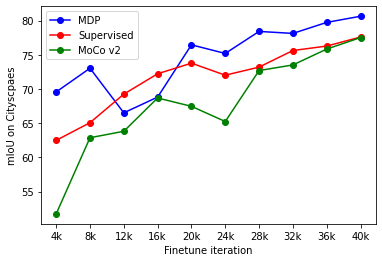}
        \label{fig:main_city}
        \vspace{-0.3cm}
    \end{minipage}
    }
    \vspace{-0.2cm}
    \caption{Comparisons of MDP, MoCo-V2, and supervised ImageNet pretraining on a) Pascal VOC, b) ADE20k and c) Cityscapes at different fine-tuning iterations. MDP consistently outperforms the other two models throughout the training procedure.}
\label{results_main}
\end{figure*}

\subsection{Ablation Study} 

\begin{table}[t]
\centering
\renewcommand\arraystretch{1.15}
\caption{
Ablation studies on hyper-parameters and the type of memory bank.
}\label{t1} 
    \begin{tabular}{@{} c  c c c c @{}}
      \toprule
    Method  & \multicolumn{2}{c}{Memory Bank}  & Temperatue $\tau$ &  mIoU  \\
       & Type & Size& & \\
      \midrule
      pixel-to-pixel & & & 0.07  &68.05\\
      MDP & Region & 4096 & 0.07 & 68.69\\
      MDP & Class & 4096 & 0.07 & \textbf{69.72}\\
      \midrule
      MDP  & Class & 1024 & 0.07  & 69.61\\
      MDP & Class & 4096 & 0.07  & 69.72\\
      MDP & Class & 10560 & 0.07 & \textbf{69.8}\\
       \midrule
      MDP & Class & 4096 & 0.07  & \textbf{69.72}\\ 
      MDP & Class & 4096 & 0.3  & 69.01\\
      MDP & Class & 4096 & 0.5  & 68.65\\
      \bottomrule
    \end{tabular}
\end{table}

\begin{table}[t]
\centering
\caption{
Comparisons of different cross-dataset mixing strategy. 
}\label{t2} 
\renewcommand\arraystretch{1.15}
    \begin{tabular}{@{} c |c c| c c |c @{}}
      \toprule
    Method  & \multicolumn{2}{c|}{Pretrained Dataset}  & \multicolumn{2}{c|}{ Augmentation Type} &  mIoU  \\
       & VOC & ADE20K& Region-level& Pixel-level & \\
      \hline
      \multirow{3}{*}{MDP} & \checkmark & & & &69.72\\
       & \checkmark & &  \checkmark& &\textbf{70.92}\\
       & \checkmark & &  & \checkmark&68.40\\
      \hline
      \multirow{5}{*}{MDP}& \checkmark &\checkmark & & &71.98\\
      & \checkmark &\checkmark & \checkmark
&  &72.66\\
      & \checkmark &\checkmark&  & \checkmark& 73.00\\
      & \checkmark &\checkmark& \checkmark & \checkmark&\textbf{73.32}\\
      \bottomrule
    \end{tabular}
\end{table}

\begin{table}[t]
\centering
\caption{
Results of utilizing cross-class sparse coding.
}\label{t3} 
\renewcommand\arraystretch{1.15}
    \begin{tabular}{@{} c |c c |c|c c @{}}
      \toprule
    Method  & \multicolumn{2}{c|}{Pretrained Dataset} &Sparse coding &   \multicolumn{2}{c}{mIoU}  \\
       & VOC & ADE20K& & VOC & Cityscpes  \\
      \hline
    MDP &\checkmark &\checkmark & &\textbf{71.98}& 75.37\\
    MDP &\checkmark &\checkmark & \checkmark &71.84&\textbf{76.53}\\
      \bottomrule
    \end{tabular}
\end{table}
In this section, we conduct extensive ablation studies to better understand how each component affects the performance. \textbf{Unless specified, all models are pretrained over Pascal VOC for 100 epochs and evaluated on Pascal VOC for efficiency.}

\textbf{Hyperparameter analysis.}
Table \ref{t1} studies the influence of the temperature $\tau$ and the memory bank size $N_b$. Note that the actual size of the class prototype in our implementation is $N_b/bs$.
We conclude that the proposed algorithm is relatively robust to memory bank size $N_b$ when sufficient samples are stored in the memory bank.
We also find that the temperature  $\tau=0.07$ brings better performance under supervised pixel level contrastive learning setting, which is different from self-supervised learning where $\tau$ is usually larger ($\tau=0.2$ in MoCo-v2). The reason is that $\tau$ controls the strength for contrastive learning, with smaller $\tau$ indicating stronger penalizing for compactness and separability. It makes sense that pixel level contrastive learning gets better results for small $\tau$ since it is guided by ground truth labels.

\textbf{Effects of class prototype.} 
Table \ref{t1} reveals the effectiveness of class prototype. Our method surpasses pixel-to-pixel contrastive learning baseline (r.f. Sec. \ref{sec:baseline}) by 1.67\%. Compared with the region-based prototype (r.f. Sec. \ref{sec:prototype}) which stores the embedding of each image by class, our method obtains $1.03\%$ performance gain (68.69\% to 69.72\%).
This is mainly due to that our method can obtain more representative embedding and has the ability to resist the class imbalance problem.

\textbf{Effects of cross-dataset mixing.} 
Table \ref{t2} inspects the influence of cross-dataset mixing. It can be seen that when both pretraining and evaluation are performed on Pascal VOC, only region-level mixing can bring gains. We think this is because pixel-level mixing changes the distribution of labels, which is not beneficial when using the same, single dataset for pretraining. However, when extended to multiple datasets, both mixing methods bring significant performance improvements (0.69\% gains for region-level mixing and 1.02\% gains for pixel-level mixing), and combining them brings another improvement (73.32\%).

\textbf{Effects of cross-class sparse coding.}
We also diagnose the effectiveness of cross-class sparse coding. As shown in Table \ref{t3}, cross-class sparse coding brings 1.16\% performance improvement on Cityscapes, from 75.37\% to 76.53\% but achieves no performance gain on Pascal VOC dataset (71.84\% compared to 71.98\%). We think this is because both pretraining and fine-tuning make use of Pascal VOC data. In this case, it may be more effective to directly separate the label space of Pascal VOC from the label space of other datasets. While sparse coding is suitable for better transferability. 

\textbf{Does MDP obtain more discriminative features?} 
Fig. \ref{results_main} compares the evaluation results of MDP and ImageNet pretraining at different fine-tuning iterations over three datasets. It can be seen that the performance of MDP is substantially higher than ImageNet pretrained model. It should be noted that at the beginning of fine-tuning, MDP has achieved far better performance, even for Cityscapes that the model does not see during the pretraining stage, which proves that MDP can obtain more discriminative features due to pixel level learning. MoCo-v2 suffers the worst initial performance, which also indicates that the instance-level contrast learning cannot handle pixel level semantic segmentation tasks well.

\section{Conclusion}
\label{sec:con}
This paper proposed a multi-dataset pretraining framework for semantic segmentation, which can efficiently make use of the off-the-shelf annotations to construct a better and more general pretrained model. The main contributions are two folds. First, we propose a pixel-to-class prototype contrastive loss to effectively model intra-class compactness and inter-class separability of multiple datasets regardless of their taxonomy labels. Second, we extend the pixel level embeddings via cross dataset mixing and pixel-to-class sparse coding for better transferability. Our method consistently outperforms the pretrained models over ImageNet on several widely used benchmarks, while using much fewer samples for pretraining. Although MDP only makes use of fully labeled data for pretraining currently, our method is able to conveniently extended to the semi-supervised scenario, which remains to be explored in the future work.

\bibliographystyle{plain}
\bibliography{neurips_2021}


\end{document}